\def\year{2022}\relax
\documentclass[letterpaper]{article} 
\usepackage{aaai22}  
\usepackage{times}  
\usepackage{helvet}  
\usepackage{courier}  
\usepackage[hyphens]{url}  
\usepackage{graphicx} 
\usepackage{natbib}  
\usepackage{caption} 
\DeclareCaptionStyle{ruled}{labelfont=normalfont,labelsep=colon,strut=off} 
\usepackage{algorithm}
\usepackage{algorithmic}

\usepackage{comment}                           
\usepackage{multirow}
\usepackage{todonotes}                          
\usepackage{subcaption}                          
\usepackage{blindtext}     
\usepackage{varioref}
\usepackage{arydshln}
\usepackage{cleveref}
\crefname{section}{s}{ss}
\crefname{section}{s}{ss}
\crefname{table}{Table}{}
\crefname{figure}{Fig.}{}
\crefname{algorithm}{Alg.}{}
\crefname{ALC@unique}{Line}{Lines}
\crefname{equation}{Eq.}{}
\crefname{appendix}{Appendix}{}
\crefformat{section}{\S#2#1#3}

\usepackage{newfloat}
\usepackage{listings}
\floatstyle{ruled}
\newfloat{listing}{tb}{lst}{}
\floatname{listing}{Listing}

\title{Transferring Semantic Knowledge Into Language Encoders}
\author{
    Mohammad Umair, %
    Francis Ferraro
}
\affiliations{
Department of Computer Science and Electrical Engineering \\
University of Maryland, Baltimore County \\
\texttt{ferraro@umbc.edu}
}

\usepackage{bibentry}

\begin{document}
\maketitle


\begin{abstract}
We introduce semantic form mid-tuning, an approach for transferring semantic knowledge from semantic meaning representations into transformer-based language encoders. In mid-tuning, we learn to align the text of general sentences---not tied to any particular inference task---and structured semantic representations of those sentences. Our approach does not require gold annotated semantic representations. Instead, it makes use of automatically generated semantic representations, such as from off-the-shelf PropBank and FrameNet semantic parsers. We show that this alignment can be learned implicitly via classification or directly via triplet loss. Our method yields language encoders that demonstrate improved predictive performance across inference, reading comprehension, textual similarity, and other semantic tasks drawn from the GLUE, SuperGLUE, and SentEval benchmarks. We evaluate our approach on three popular baseline models, where our experimental results and analysis concludes that current pre-trained language models can further benefit from structured semantic frames with the proposed mid-tuning method, as they inject additional task-agnostic knowledge to the encoder, improving the generated embeddings as well as the linguistic properties of the given model, as evident from improvements on a popular sentence embedding toolkit and a variety of probing tasks.
\end{abstract}


\section{Introduction}

The basic transformer architecture, first proposed by \citet{vaswani2017attention}, has quickly replaced recurrent cells as the ``go-to'' neural architecture. Pre-trained models like BERT~\cite{devlin-etal-2019-bert} have shown how to build effective language encoders via a set of two, simple pre-training tasks: an indirect language modeling task (next sentence prediction), and a direct language modeling task (masked language modeling). Those two pre-training tasks have been shown to be effective in leveraging implicit language understanding to better address semantic-oriented inference tasks.

\smallskip

However, while BERT may be effective at encoding contextual meaning implicitly, we wonder: how can we incorporate \textit{general} meaning and semantics more directly into BERT, but in as straightforward a way as possible? %
Broadly, our aim is to incorporate semantics from structured meaning representations in ways that can improve the encoded representation of surface (lexical), syntactic (rules of language), and semantic (meaning and logistics of sentence) information. %
While we wish to \textit{directly} and \textit{explicitly} inject knowledge into the encoding process, we are not looking for a new transformer architecture. Nor are we looking for a resulting encoder that depends on having access to knowledge in downstream uses. Rather, we are looking for the adaptation of an existing encoder that enables broad-domain (i.e., end-task agnostic) semantic representations to yield improvements in downstream tasks. 

To accomplish this, we introduce a method that we call ``semantic mid-tuning'', where we incorporate automatically extracted semantic representations into a pre-training-like objective. Our approach guides a language encoder to incorporate a richer formalism of semantics by synchronously learning to embed structured meaning representations and then using those learned embeddings in an indirect language modeling pre-training task. We demonstrate that our semantic mid-tuning approach is flexible, by experimenting with two common semantic representations that operate at different levels of granularity: a rather coarse-grained PropBank-style~\citep{palmer-etal-2005-proposition} semantic parse, and a fine-grained FrameNet~\citep{baker1998berkeley} semantic parse. Our goal is not to get SOTA, but instead demonstrate consistent improvements by using our alignment approach. When evaluated on the well-known GLUE,  SuperGLUE, and SentEval benchmarks, we find that our approach improves a number of inference, semantic similarity, and sentiment analysis tasks across these datasets. The core contributions of our method are: (1) we demonstrate that sentence-based structured semantic representations can be incorporated into transformer-based learning in a way that adds additional semantic knowledge to the transformer; (2) we demonstrate that this task-agnostic knowledge transfer improves downstream semantic inference task performance; (3) we demonstrate that this additional mid-tuning step improves the linguistic properties of sentence embedding models. Upon publication, our code and data will be made publicly available. 



\section{Method}


The approach we use is inspired by Sentence-BERT~\cite[SBERT]{reimers2019sentence}, a popular approach for learning improved language encoders. SBERT fine-tunes BERT (or generally, any language encoder) using a parallel network architecture to generate efficient sentence embeddings: two copies of a base encoding function take two input sequences, embed them, and compare the embeddings to determine how similar they are. The embedding function is then (further) trained such that inputs that are related to each other yield similar embeddings. In our case, one input sequence is the text of a sentence, and the other is the structured meaning representation. 

\subsection{Datasets}  
\label{datasets}

We define a \textit{semantic form} as a structured semantic representation, such as a semantic parse. %
To show the impact that providing even a small amount of semantic forms can have, we take approximately 250,000 sentences from an existing English Wikipedia archive~\cite{ferraro2014concretely}. We chose Wikipedia as its more general encyclopedic knowledge has been shown to be a useful resource for broader knowledge extraction~\cite{ponzetto2007deriving}. This is a \textit{very} small amount of extra task-agnostic data, amounting to only 0.5\% of English Wikipedia.\footnote{While the SRL parses we use are automatically obtained, we acknowledge that our training approach relies on resources that are not available for all languages.} %
For our core experiments, we obtain a PropBank-style (``PB'') semantic parse from \citet{zhang2019semantics} which provides methods to generate semantically-rich PropBank annotations using the \citet{he2017deep} approach. 
From a sentence, the meaning representations contain extracted frames, roles, triggering lexical predicates, and lexical arguments (role fillers). We provide the structured semantic forms to the semantic form encoder and mid-tuning adjusts the encoder to embed this structure.





\subsection{Learning and Aligning Encoders}
\label{sec:learning}

We learn a sentence encoder $E_1$ and a semantic form encoder $E_2$. Both are initialized with off-the-shelf transformers. We encode the sentence with $E_1$ and the semantic form with $E_2$.\footnote{To get a fixed-size vector/embedding of each, a mean-pooling method is used where each pair is passed to the encoding model separately, and a fixed-sized embedding is computed by averaging the contextualized output vectors for all tokens.} These encoders produce initial embeddings of the sentence $s_a$ (respectively, semantic form $r$), which are then aligned to be ``close.'' During the training, we prepend special tokens to each input pair which helps us further differentiate between a sentence and meaning representation. \footnote{   
We take the idea of adding a special token id/tag to each input sentence in our mid-tuning dataset, from \citet{gao2020complementing} and \citet{liu2020multilingual}. It helps the model to differentiate between sentence $s_a$ and semantic form $r$). We use the token [“\_EN\_”] for sentence, and [“\_SRLMR\_”] for the PropBank representations. This is quite simple in terms of code but impacts the overall scores nicely. 
} %
Once trained, we use the sentence encoder $E_1$ for fine-tuning on different benchmarks. %

A number of different objectives can train this parallel architecture. We use cross-entropy and triplet loss, which we argue are forms of semantic manifold alignment: the sentence and semantic embeddings that a non-mid-tuned transformer produces potentially live in different spaces, where our mid-tuning aligns the embeddings, and in doing so transfers semantic meaning. 
Both rely on negative sampling to get training negative instances. %
We use multiple deletion functions to filter parts of the original representation $r$ by either removing roles and/or their fillers, or swapping the fillers of different roles. 

\paragraph{Implicit Alignment via Cross-Entropy}
Cross-entropy (classification) uses $s_a$ and $r$ as features in predicting whether the semantic form that produced $r$ is the correct (extracted) form for the sentence that produced $s_a$. Here, cross-entropy \textit{implicitly} aligns $s_a$ and $r$, which mid-tunes the encoders $E_1$ and $E_2$. 

We take the sentence-representation pairs and create a balanced corpus for binary classification: we predict ``yes'' if the meaning representation and sentence are paired, and ``no'' if not. This classification task can be considered a form of semantic parse recognition without partial credit. The false instances are generated via negative sampling: given a sentence, an incorrect representation is randomly assigned to it, so that the trained model is unbiased and understands both correct and incorrect cases. %

\paragraph{Explicit Alignment via Triplet Loss}
Triplet loss uses a contrastive loss to mid-tune $E_1$ and $E_2$. We encode both the extracted meaning representation ($r_+$) and some other meaning representation ($r_-$). We treat $s_a$ as the ``anchor'', and using a distance $d$, triplet loss learns $E_1$ and $E_2$ so that $d(s_a, r_+)$ is small while $d(s_a, r_-)$ is large. We use Euclidean distance for $d$, so our objective is: 
\begin{equation}
    \max\;(\:||s_a-r_+||^2_2-||s_a-r_-||^2_2+\epsilon,\;0\:).
\end{equation}
\noindent In doing so, triplet loss \textit{explicitly} aligns $s$ and $r$. %

\section{Experiments}

To measure the effectiveness of our approach, we perform a number of experiments on three popular NLP language modeling benchmarks. A pre-trained language model can be applied to a given supervised downstream task in two ways. i.e., feature extraction or fine-tuning. In feature extraction, the pre-trained model weights are kept frozen and the output representations are fed to another model. Whereas, in fine-tuning, the pre-trained model itself is trained on the target task, which allows faster convergence compared to the random initialization, as the pre-trained model act as a starting point for the target model. The benchmarks GLUE and SuperGLUE measures the performance of fine-tuning while the SentEval toolkit measures the effectiveness of the feature extraction approach. These experiments help us understand the strength and weakness of our method, and analyze the linguistic knowledge it improves within transformers. We do not compare our mid-tuning results with previous embedding approaches like InferSent \cite{conneau2017supervised} or Universal Sentence Encoder \cite{cer2018universal}, as such comparisons would not be fair. Our goal is to demonstrate that semantic mid-tuning improves a given baseline transformer model, not to best all previous approaches. 

\begin{table*}
{\fontsize{8.5}{13} \selectfont
\centering 
\begin{tabular}{l|rr|rrr|rrrrr}      


\hline \hline \textbf{Model} &\textbf{Len}          
&\textbf{WC}        
&\textbf{Depth}    
&\textbf{TConst}    
&\textbf{BShift}    
&\textbf{Tense}     
&\textbf{SbjNum}   
&\textbf{ObjNum}    
&\textbf{OMO}       
&\textbf{CoInv}     
\\ \hline \hline



BERT-base   & 68.05 & 50.15 
                 & 34.65 & 75.9
                 & 86.41 & 88.81 & 83.36 & 78.56 & 64.87 & 74.32 \\ 

\hdashline

+ PB + Triplet ${\dagger}$    & \textbf{76.19} &            \textbf{54.44} & \textbf{36.75} &               \textbf{79.64} & \textbf{89.15} &                \textbf{89.33} & \textbf{86.15} &           79.81 & \textbf{65.08} & \textbf{75.13} \\


+ PB + Cls.   & 73.68 & 54.04 
                 & 32.09 & 56.1
                 & 64.86 & 82.43 & 80.76 & \textbf{80.62} & 55.65 & 54.13\\


 \hline \hline

SBERT-base   & 75.55 & 58.91 
                 & \textbf{35.66} & 61.49
                 & 77.93 & \textbf{87.32} & 79.76 & \textbf{78.40} & 62.85 & \textbf{65.33} \\

\hdashline


 + PB + Triplet  ${\dagger}$   & \textbf{76.33} & \textbf{62.25}
                 & 35.33 & \textbf{62.61}
                 & \textbf{79.24} & 86.12 & \textbf{80.67} & 78.35 & \textbf{64.05} & 65.11 \\

 + PB + Cls.   & 73.73 & 58.08 
                 & 32.26 & 53.85
                 & 58.34 & 77.54 & 76.43 & 77.19 & 53.85 & 52.72 \\

                 
    \hline 
XLNet-base   & 64.01 & 24.63 
                 & 32.18 & 79.55
                 & 68.2 & 85.34 & 86.35 & 73.74 & 55.31 & 67.59 \\ 

\hdashline

 + PB + Cls. ${\dagger}$ & 66.13 &                             \textbf{46.11} 
                 & \textbf{36.61} & 78.47 
                 & 70.81 & \textbf{86.6} & \textbf{91.17} & \textbf{85.9} & \textbf{58.68} & 66.5 \\


 + PB + Triplet   & \textbf{70.51} & 22.69 
                 & 35.22 & \textbf{79.65}
                 & \textbf{74.0} & 85.74 & 88.64  & 83.23 & 58.36 & \textbf{67.67} \\


\hline
    \end{tabular}   
    \caption[SentEval Probing Results]{Results on \textbf{SentEval Probing tasks}. Probing datasets evaluate the linguistic properties of a model. We present the test set scores here. ${\dagger}$ presents the best performing model against each baseline model. Our mid-tuning models outperform all probing tasks in comparison to each baseline model.}
\label{senteval_probing}

}
\end{table*}

We acknowledge other approaches similar to semantic mid-tuning have been proposed, such as \citet{phang2018sentence} and \citet{arase2019transfer}. We do not compare with them due to two critical differences: the form of the knowledge and the intermediate task relevance. First, those work on paired sentences rather than sentences and structured representations. Second, \citet{phang2018sentence} use intermediate tasks relevant to the particular downstream inference tasks; we don’t. Comparing against them would present additional factors that we would need to control for in order to have a fair comparison.

\subsubsection{Experimental Settings}

The mid-tuning models are trained with 1 English Wikipedia file of 250000 sentences, with a batch size of 64. Early results showed 1 epoch to be sufficient. On average, it takes about 2-4 hours to train on a single RTX 8000 GPU. Our core experiments include the results of mid-tuning models trained on PropBank (``PB'') representations, with three different {\bf\textlangle Base\textrangle} models: BERT-base~\cite{devlin-etal-2019-bert} for a basic conditional transformer method, SBERT~\cite{reimers2019sentence} to show how mid-tuning works with additional (task-relevant) training, and XLNet \cite{yang2019xlnet} to demonstrate mid-tuning applied to generative LMs. BERT-base and SBERT-base are uncased while XLNet is cased. We use the following mid-tuning strategies: 
\begin{itemize}
\small

\label{model_names_1}
    \itemsep0em  
    \item \textbf{\textlangle Base\textrangle+PB+Cls.:} Base model trained on PropBank SRL forms via cross entropy/classification (implicit alignment)
    \item \textbf{\textlangle Base\textrangle+PB+Triplet:} Base model trained on PropBank SRL forms via triplet loss (explicit alignment)
\end{itemize}

\subsubsection{Evaluation Sets and Goals}
We emphasize that our goal is \textbf{not} to achieve state-of-the-art results on the downstream tasks. Rather, it is to demonstrate the ability of semantic mid-tuning to impart semantic knowledge into a given \textlangle Base\textrangle{} model. %
To that end, we evaluate on three common NLP datasets: SentEval, GLUE, and SuperGLUE.

SentEval \cite{conneau2018senteval} is a toolkit to evaluate the effectiveness and generalization of sentence representations/embeddings, on a set of different downstream classification, inference, similarity, and probing tasks. The probing tasks examine linguistic information at the \textbf{surface-level} (how well embeddings encode surface knowledge that does not require linguistic information); the \textbf{syntactic-level} (how well the embeddings encode the grammatical structure of a sentence; and the \textbf{semantic-level} (how well the embeddings encode the meaning and logistics behind the sentences). For evaluating on SentEval, we use the scripts provided by SBERT-WK \cite{wang2020sbert}. We use the ``CLS'' embedding method for BERT-base and XLNet-base, while ``Ave\_last\_hidden'' for the SBERT-base model and replicate the results with the original paper. For fine-tuning parameters, we use Max\_Sequence\_Length of 128, Batch\_Size of 64, 4 number of epochs, using Adam Optimizer with 10-fold cross-validation for all tasks, which are the same settings as SBERT-WK. SentEval has six polarity assessment and three entailment/inference tasks (supervised downstream tasks); seven semantic textual similarity (STS) tasks; and ten probing tasks.

GLUE \cite{wang2018glue} is a set of nine diverse English understanding tasks that are often used to evaluate sentence representation models. The GLUE tasks require the model to be able to reason, do inference, and have a comprehensive understanding of natural language. For fine-tuning, we use the standard, publicly available scripts from  \url{https://github.com/huggingface/transformers}{huggingface} transformers with their default parameter settings of a per\_gpu batch size of 8, Max\_Sequence\_Length of 128, learning rate 2e-5, with 3 training epochs. Our results on GLUE varies from \citet{devlin-etal-2019-bert}'s scores for their baseline model since they use a batch size of 32, fine-tune for 3 epochs, and use different learning rates (ranging from 5e-5, 4e-5, 3e-5, and 2e-5) and report the best scores. 

\begin{table*}
{\fontsize{9}{13} \selectfont
\centering 
\begin{tabular}{l|rrrrr|r|r}      
\hline \hline \textbf{Model} &\textbf{STS12}              
&\textbf{STS13}                       
&\textbf{STS14}
&\textbf{STS15}
&\textbf{STS16}
&\textbf{STSb}
&\textbf{SICK-R}

\\ \hline \hline


BERT-base   & 0.27/0.32 & 0.22/0.23 
                 & 0.25/0.28 & 0.32/0.35
                 & 0.42/0.51 & 0.52/0.51 & 0.70/0.64 \\ 

\hdashline

 + PB + Cls. ${\dagger}$   &  \textbf{0.53/0.54}
                 & \textbf{0.55/0.56} 
                 & \textbf{0.60/0.60} & \textbf{0.62/0.63}
                 & \textbf{0.61/0.62} & \textbf{0.65/0.64} & \textbf{0.81/0.75} \\


 + PB + Triplet    & 0.28/0.33 & 0.24/0.24 
                 & 0.27/0.28 & 0.34/0.36
                 & 0.43/0.51 & 0.58/0.57 & 0.75/0.69 \\     

 \hline \hline 

SBERT-base & 0.64/0.63 & 0.67/0.69             
                 & \textbf{0.73/0.73} & 0.74/0.75        
                 & 0.70/0.73 & 0.74/0.74 & 0.84/0.79 \\

\hdashline

 + PB + Triplet ${\dagger}$   
                & \textbf{0.67/0.66} &                  \textbf{0.75/0.74} & 0.71/0.72 &      \textbf{0.76/0.78} & \textbf{0.73/0.76} & \textbf{0.77/0.77} &  0.84/0.79 \\



 + PB + Cls.    & 0.49/0.53 &                                              0.54/0.56 & 0.63/0.63 & 0.63/0.64
                 & 0.57/0.60 & 0.71/0.71 & 0.83/0.76 \\


    \hline \hline

XLNet-base   & 0.18/0.23 & 0.02/0.02 
                 & 0.07/0.12 & 0.08/0.13
                 & 0.20/0.25 & 0.37/0.39 & 0.48/0.49 \\ 

\hdashline

 + PB + Cls. ${\dagger}$  & \textbf{0.30/0.37} & \textbf{0.25/0.33} 
                 & \textbf{0.31/0.36} & \textbf{0.33/0.35}
                 & \textbf{0.34/0.40} & \textbf{0.43/0.41} & \textbf{0.67/0.63} \\


 + PB + Triplet   & 0.08/0.20 & 0.08/0.13 
                 & 0.09/0.16 & 0.09/0.17 
                & 0.17/0.28 & 0.38/0.39 & 0.61/0.59 \\


\hline 

    \end{tabular}   
    \caption{Results on the test sets for the \textbf{SentEval} Textual Similarity tasks. We report Pearson and Spearman correlation for all tasks. Best results are presented in bold faces while dagger ${\dagger}$ represents the best performing model over the baseline. } 
\label{senteval_2}
}
\end{table*}   

SuperGLUE \cite{wang2019superglue} is an updated version of the GLUE benchmark, with a set of more challenging language understanding tasks like question answering, entailment, and reasoning. The scripts we use are taken from \url{https://github.com/huggingface/transformers}{huggingface transformers} GitHub repository and modified for the selected tasks. Our results vary from those reported by \citet{wang2019superglue} for the BERT baseline, since they use different parameter configurations (a learning rate of 10e-5 and fine-tune for 10 epochs) on a different baseline model (they use large cased, we use base uncased) for all tasks. For fine-tuning BERT models, we use default settings of per\_GPU\_batch\_size of 8, learning rate 5e-5, warm-up ratio of 0.06, gradient accumulation steps of 8, and a weight decay of 0.01.

\subsection{Mid-Tuning Improves Linguistic Encodings}
\label{subsec:probing-subsection}        

In our first round of experiments, we analyze what type of information the model stores and evaluate them based on predicate-argument structure, logic/semantics, knowledge, and common sense. We do this via the SentEval probing tasks that evaluate the linguistic properties encoded in sentence embeddings. As seen in Table~\ref{senteval_probing}, incorporating semantic representations outperforms all probing tasks over the baseline models, which continues to suggest that these structured meaning representations improve not only the surface (lexical) but syntactic (rules of language) and semantic (meaning and logistics of sentence) information as well.

Looking into the probing tasks, we note that \citet[p.~2]{conneau2018you} concludes that the \textbf{Surface} tasks (Len, WC) ``can be solved by simply looking at tokens in the input sentences, and do not require linguistic knowledge.'' While it would be reasonable then to assume that lexical-only methods (i.e., the base models) would have an advantage, we see that mid-tuning provides non-trivial improvements. 
The \textbf{Syntactic} tasks (Depth, TConst, BShift) are based on the syntactic structure of the sentences. Notably 
\textit{TConst} is designed to reflect encoded and clustered syntactic structures, like (latent) constituents, while \textit{BShift} addresses lexical correspondence. %
Finally, \textbf{Semantic} tasks (Tense, SbjNum, ObjNum, OMO, CoInv) also ``rely on syntactic structure, but they further require some understanding of what a sentence denotes'' \Citet[p.~3]{conneau2018you}. 
When we consider that one goal of semantic representation is to be able to use a representation that abstracts away from the lexical and even syntactic choices one can make, the results suggest that mid-tuning enables that abstraction to be captured. %

The results on probing tasks show that the linguistic properties of an encoder can be improved when structured frames are aligned with our method, and we see improvements with almost all probings tasks over the baselines models we evaluate our proposed approach upon. They support our argument that mid-tuning adds additional general semantic knowledge to the encoders.

\subsection{Mid-Tuning Improves Semantic Text Similarity}

Next, we evaluate our approach on 7 different semantic similarity tasks, presented in Table~\ref{senteval_2}. Mid-tuning greatly improves all STS tasks with a great margin (some with 20-30 point absolute improvement). Implicit alignment with cross-entropy proves to be very helpful for STS tasks with both BERT and XLNet models. We note significant improvements over the SBERT model too, which itself has very strong performance on these STS tasks. While the off-the-shelf SBERT-base is already trained via a task-relevant classification objective, it benefits further from our task-agnostic triplet alignment mid-tuning. %

\begin{table*}
{\fontsize{9}{13} \selectfont
\centering 
\begin{tabular}{l|rrrrrr|rrr}      

\hline 
&
\multicolumn{6}{c}{Polarity Assessment Tasks} &
\multicolumn{3}{|c}{Inference Tasks}\\
\hline 

\hline \hline \textbf{Model} &\textbf{MR}
&\textbf{CR}
&\textbf{SUBJ}
&\textbf{MPQA}
&\textbf{SST2}
&\textbf{TREC}
&\textbf{MRPC}
&\textbf{SICK-E}
&\textbf{RTE}

\\ \hline \hline



BERT-base   & \textbf{82.19} & \textbf{87.58} 
                 & \textbf{95.52} & 88.52
                 & \textbf{86.88} & \textbf{93.8} & 71.25 & 73.84 & 56.32\\ 

\hdashline

 + PB + Triplet ${\dagger}$   & 82.12 & 87.05 
                 & 95.22 & 88.17
                 & 86.71 & 92.2 & 73.68 & 76.4 & \textbf{62.45}\\


 + PB + Cls.   & 75.45 & 81.19 
                 & 90.28 & \textbf{89.05}
                 & 80.51 & 85.4 & \textbf{74.2} & \textbf{79.76} & 59.57\\


 \hline \hline

SBERT-base  & \textbf{82.49} & \textbf{88.8} 
                 & \textbf{94.25} & \textbf{90.2}
                 & \textbf{88.36} & 86.4 & 75.48 & 82.28 & 59.57 \\

\hdashline


 + PB + Triplet ${\dagger}$ & 82.27 & 88.72     
                 & 94.11 & 89.93        
                 & 87.92 & \textbf{87.4} & \textbf{75.77} & 82.57 & \textbf{59.93}\\


 + PB + Cls.  & 73.2 & 80.27 
                 & 89.36 & 88.82             
                 & 76.94 & 84.6 & 72.75 & \textbf{83.3} & 56.68\\


    \hline \hline

XLNet-base   & 78.4 & 83.23 
                 & 91.38 & 87.26
                 & 81.82 & 90.2 & 66.49 & 56.69 & 50.28\\ 

\hdashline

 + PB + Cls. ${\dagger}$  & \textbf{80.41} & \textbf{87.29} 
                 & \textbf{92.54} & \textbf{89.24} 
                 & \textbf{85.94} & \textbf{91.4} & \textbf{72.93} & \textbf{76.38} & \textbf{58.43}\\

 + PB + Triplet   & 78.84 & 85.01 
                 & 90.6 & 84.72 
                 & 84.73 & 87.0 & 68.87 
                 & 71.02 & 52.53\\



\hline

    \end{tabular}   
    \caption[SentEval Downstream Results]{Results on \textbf{SentEval} Supervised Downstream tasks. RTE, SICK-E, and MRPC fall in the inference category while the rest are classification tasks. We present the test set scores here. RTE tasks have been manually added here by modifying the scripts for MRPC. The best results are shown in bold faces, whereas $\dagger$ represents the best performing model.  }
\label{senteval_1}
}
\end{table*}   

\subsection{Mid-Tuning Provides Consistent Downstream Inference Improvements}

Overall we observe notable impact of our mid-tuning method when applied to supervised downstream tasks. %
First, we look at the supervised downstream SentEval tasks presented in Table~\ref{senteval_1}; then we look at the downstream GLUE and SuperGLUE performance in \cref{glue_superglue_table}. 

For SentEval, we divide the results into ``Polarity Assessment'' and ``Inference'' tasks. Mid-tuning was competitive with the baseline methods on the polarity assessment tasks, often coming within 0.1 point of the baseline. On the other hand, \textbf{all} mid-tuning models outperformed \textbf{all three} baseline models for \textbf{all inference tasks}. In contrast to the near competitive performance on the polarity tasks, mid-tuning consistently resulted in multiple point improvements, with some notably large improvements: mid-tuning with triplet loss brought RTE from 56.32 to 62.45 for BERT-base, and from 50.28 to 58.43 for XLNet-base. Additionally, XLNet is the only model where mid-tuning also outperforms all polarity assessment tasks, whereas BERT and SBERT does not show any improvement, but remain equally effective. 

For space and readability, we present select GLUE and SuperGLUE results in \cref{glue_table} and \cref{superglue_table}, respectively. %
For both BERT and SBERT, mid-tuning outperforms a number of tasks on GLUE when compared to the baseline model. While the margin of improvement may not be large, we note that mid-tuning retains accuracy on the rest of the tasks. In contrast, we see good improvements against XLNet-base \cite{yang2019xlnet} model, where it outperforms 6 out of 9 GLUE tasks with a decent margin. 
Note that both BERT and SBERT are autoencoders, while XLNet is autoregressive: this suggests that mid-tuning knowledge transfer can overcome downstream limitations in autoregressive models. %


Specifically for SuperGLUE, overall we see that mid-tuning provides consistent improvements over the base models. While neither triplet-based mid-tuning nor classification-based mid-tuning dominates the other, both provide consistent improvement over BERT-base and SBERT-base. The main exceptions occur with SBERT-base on COPA and BoolQ. However, we note that SBERT-base has already undergone \textit{additional} training that is relevant to the inference-related COPA and BoolQ tasks. In contrast, our mid-tuning approach is agnostic to the downstream tasks. We see that using task-relevant data to do additional training of the encoders can be beneficial, but in the vast majority of tested cases, it does not conflict with semantic mid-tuning. %
This suggests that the two approaches can be used jointly and beneficially.

Consistent with the SentEval inference-style tasks, we continue to see overall improvement on GLUE and SuperGLUE inference-style tasks. Looking at the GLUE benchmark results from Table~\ref{glue_table}, \textit{RTE} and \textit{WNLI} are inference tasks, whereas, from SuperGLUE Table~\ref{superglue_table}, the mid-tuning models greatly improve \textit{CB}, a 3-class entailment task; as well as \textit{RTE} and \textit{SICK-E} from SentEval Table~\ref{senteval_1}, all of which are inference tasks. The effect of mid-tuning alignment on inference/entailment tasks is noteworthy as in some cases, we see notable improvements over the baseline models (10 -- 20 points absolute improvement) with XLNet on SentEval, and \textit{CB} from SuperGLUE.





\begin{table*}
\begin{subtable}{.55\textwidth}
\centering 
\resizebox{.99\columnwidth}{!}{
\begin{tabular}{l|r|r|r|r|r|r|r|r|r}      
\hline \hline \textbf{Model} &\textbf{MRPC}              
&\textbf{RTE}                       
&\textbf{SST-2}
&\textbf{COLA}
&\textbf{QNLI}
&\textbf{STS-B}
&\textbf{QQP}
&\textbf{WNLI}
&\textbf{MNLI} \\ \hline \hline

BERT-base   & 0.86 & 0.67 
            & 0.92 & 0.58 
            & 0.91 & 0.89 & 0.91 & 0.56 & 0.84\\ 

\hdashline

+ PB + Triplet ${\dagger}$   & \textbf{0.87} & \textbf{0.69} 
                & 0.92 & \textbf{0.60} 
                & 0.91 & \textbf{0.90} & 0.91 & 0.56 & \textbf{0.85}\\      

+ PB + Cls.   & 0.85 & 0.67 
            & 0.92 & 0.59 
            & 0.91 & 0.90 & 0.90 & 0.53 & 0.85\\


\hline 
SBERT-base & 0.87 & \textbf{0.72} 
            & \textbf{0.93} & 0.54
            & 0.91 & 0.89 & 0.91 & 0.37 & 0.84 \\

\hdashline

+ PB + Triplet ${\dagger}$  & 0.87 & 0.71 
                 & 0.92 & \textbf{0.57}
                 & 0.91 & 0.89 & 0.91 & \textbf{0.49} & 0.84 \\

+ PB + Cls. & 0.82 & 0.66 
            & 0.92 & 0.53
            & 0.91 & 0.88 & 0.90 & 0.32 & 0.84\\

                 

    \hline 

XLNet-Base   & 0.87 & 0.58 
                 & 0.93 & 0.42
                 & 0.90 & 0.87 & 0.87 & 0.39 & 0.84 \\

\hdashline

+ PB + Triplet ${\dagger}$  & 0.87 & \textbf{0.64} 
                 & 0.93 & 0.45
                 & 0.89 & 0.87 & \textbf{0.91} & \textbf{0.56} & \textbf{0.86} \\

+ PB + Cls.   & 0.87 & 0.57 
                 & 0.93 & \textbf{0.48}
                 & \textbf{0.91} & 0.87 & 0.90 & 0.26 & 0.86 \\



\hline 
    
    \end{tabular}  
    }
    \caption[GLUE Benchmark Results]{\textbf{GLUE} results. The scores metric is Mathews's Correlation for COLA, Pearson-Spearman for STS-B, and accuracy for the rest of the tasks.}
    \label{glue_table} 
    \end{subtable}
    ~
\begin{subtable}{.43\textwidth}
\centering 
\resizebox{.98\columnwidth}{!}{
\begin{tabular}{l|r|r|r|r|r|r} 
\hline \hline \textbf{Model} &\textbf{COPA}   
        &\textbf{WSC}   
        &\textbf{CB}  
        &\textbf{BoolQ}  
        &\textbf{AX-b}  
        &\textbf{AX-g}  \\ \hline \hline
    
    BERT-base   & 0.55 & 0.52   
                     & 0.28 & 0.67   
                     & 0.58 & 0.50 \\   

\hdashline
    
     + PB + Triplet ${\dagger}$  & \textbf{0.57} & \textbf{0.55} 
             & 0.52 & 0.68    
             & 0.62 & \textbf{0.52}\\

     + PB + Cls.   & 0.52 & 0.54   
                     & \textbf{0.55} & \textbf{0.69}  
                     & \textbf{0.66} & 0.51\\

    

        \hline \hline
    SBERT-base & \textbf{0.57} & 0.52   
             & 0.55 & 0.69  
             & 0.63 & 0.49\\  

\hdashline

     + PB + Triplet ${\dagger}$  & 0.54 & \textbf{0.58}
             & \textbf{0.71} & 0.69  
             & \textbf{0.65} & \textbf{0.52}\\

     + PB + Cls. & 0.49 & 0.55   
             & 0.62 & 0.66  
             & 0.62 & 0.51\\ 
    
    
    
         \hline 

    \end{tabular} 
}
\caption[SuperGLUE Benchmark Results]{Results on the \textbf{SuperGLUE} tasks. As standard, we use Mathews's Correlation for AX-b and accuracy for the rest of the tasks.}
    \label{superglue_table} 
    \end{subtable}
    \caption{Results on \textbf{GLUE} and \textbf{SuperGLUE} benchmarks with the \textit{PropBank+Triplet} models. Dagger (${\dagger}$) is used to present these models as the best performing model against each baseline, whereas bold faces present the best scores against each baseline model on which we apply our mid-tuning approach. ``Cls.'' means mid-tuning via cross-entropy classification.
    }
    \label{glue_superglue_table}
\end{table*}   






\section{Analysis}          
\label{sec:analysis}        

\noindent This section includes a detailed study on the impact of incorporating semantic representations into our mid-tuning alignment task. We examine how our approach helps to integrate additional semantic knowledge into the pre-trained models and improve transfer learning tasks.


\subsection{Impact of incorporating semantic representations}

Overall, we have seen that mid-tuning yields language encoders that are no worse, and in many cases better, at a variety of downstream tasks and diagnostic tests. %
Our results, especially those across the inference-style and text similarity tasks, support our research aims to develop mid-tuning alignment as a way of incorporating more general meaning and knowledge into baseline language encoders. 

In particular, looking at SuperGLUE results from Table~\ref{superglue_table}, we see that mid-tuning improves \textit{WSC}, an updated version of \textit{WNLI} from GLUE; and a co-reference resolution task that requires a model to use commonsense reasoning in order to determine the correct referent of the pronoun among the list of provided choices. As \citet{wang2019superglue} mentions, BERT performs quite poorly on the \textit{WSC} task which determines the correct referent of the pronoun from among the provided choices, due to lack of data augmentation and small size of the dataset. Mid-tuning shows major improvements over both baseline models, indicating that structured semantic representations can be helpful for reasoning tasks, as SuperGLUE \cite{wang2019superglue} argues that these tasks require everyday knowledge and commonsense reasoning to solve. Our method also improves both \textit{AX-b} (Broadcoverage Diagnostics) \cite{williams2017broad} and \textit{AX-g} (Winogender Schema Diagnostics) \cite{rudinger-etal-2018-gender} tasks, both of which fall in the entailment/inference task category, where \textit{AX-b} tests the model's ability to understand lexical semantics, predicate-argument structure, and knowledge/commonsense. 

We have previously noted that mid-tuning did not always help the SBERT model. %
This was most notably seen in the polarity judgments and GLUE and SuperGLUE tasks. %
This is not too surprising, as the off-the-shelf SBERT-base is already trained with additional task-relevant information, supporting the intuitive notion that the benefit of task-agnostic knowledge can be dampened in the presence of task-relevant examples. %
Regardless, performance of mid-tuned SBERT was broadly competitive with the baseline, suggesting that the task-agnostic knowledge is generally at worst neither affirmatively beneficial nor harmful. %
However, these same mid-tuned models can greatly and affirmatively benefit tasks such as STS.



   

\subsection{Classification vs. Triplet Alignment} 


Overall, we observe that the mid-tuning alignment with the triplet objective performs better than the classification approach in most cases. We note that in case a network is sparse or includes a large number of classes in the output layer, the classification objective is less effective. In the triplet objective, a sentence is passed along with both positive and negative semantic forms as a single input pair into the same projection space, where the distributed embeddings of data points are trained in a way that contextually similar points (anchor sentence and positive representation) in the high dimensional vector space are projected close to each other and at the same time dissimilar points (anchor sentence and negative representation) are pushed far away from each other, resulting in better embeddings. Comparing SentEval classification task results, we see notable improvements with the triplet objective over the tasks where the classification approach performs poorly. i.e. \textit{MR, CR, SUBJ, and TREC}. Looking at the scores on the probing tasks, we see that the triplet approach more efficiently integrates linguistic properties in the mid-tuning models. 

However, in the case of XLNet, we see that cross-entropy is more effective and outperforms most of the SentEval tasks when compared to triplet loss. Secondly, we also note that cross-entropy alignment is highly effective for STS tasks, in case of both BERT and XLNet baseline models. In addition, we see that the implicit alignment with cross-entropy can be effective even for similarity judgments that traditionally have been learned with explicit alignment approaches unless the base model has been additionally trained with task-inspired sentences (SBERT). This may seem counter-intuitive initially, but recall that our explicit alignment is between sentences and meaning representations---not between sentences. It appears that cross-entropy is superior at providing light, \textit{effective} guidance on how to re-embed semantically similar sentences.

\begin{table}[t]
\centering 
\resizebox{.95\columnwidth}{!}{
\begin{tabular}{l|r|l|l}    
\hline \hline \textbf{Benchmark} & \textbf{Task} & \textbf{BERT-Base} & \textbf{Mid-tuning on FrameNet} \\ \hline \hline    


\multicolumn{4}{c}{\textbf{Inference Tasks}} \\
\hline

  GLUE & RTE & 0.67 & + Cls. 0.65 \\
 &  &  & + Trip. \textbf{0.71} \\
 \hdashline

 SuperGLUE & WSC & 0.52 & + Cls. \textbf{0.58} \\ 
 &  &  & + Trip. 0.53 \\ 

\hdashline

  SuperGLUE & CB & 0.28 & + Cls.\textbf{0.57} \\ 
  &  &  & + Trip. 0.46 \\ 

\hdashline

 SentEval & SICK-E & 73.84 & + Cls. \textbf{76.03} \\ 
 &  &  & + Trip. 75.73 \\

\hline
\multicolumn{4}{c}{\textbf{STS Tasks}} \\
\hline

  SentEval & STS-14 & 0.25/0.28 & + Cls. \textbf{0.43/0.44} \\ 
  &  &  & + Trip. 0.26/0.29 \\ 

\hdashline

 SentEval & SICK-R & 0.70/0.64 & + Cls. \textbf{0.75/0.68} \\ 
 &  &  & + Trip. 0.74/0.69 \\

\hline 
\end{tabular}    
}
\caption{\label{ablation-table-2} Results from using FrameNet parses to mid-tune BERT-base models. A major improvement can be seen in the selected STS and Inference tasks, when compared to the BERT-base model. }
\end{table}

\section{Effect of Semantic Frames}
\label{sec:ablation}

In this section, we perform additional experiments to examine the effect of using different types of semantic parses. We additionally provide qualitative examples of mid-tuning's impact on semantic retrieval. 

\subsubsection{Using Different Frames:}
We use the existing automatically extracted FrameNet~\cite{baker1998berkeley} parses provided by \citet{ferraro2014concretely}. 
We mid-tune the BERT baseline model on these frames and evaluate on the selected inference and STS tasks. Due to space and readability limitations, we present an illustrative excerpt of these results in Table~\ref{ablation-table-2}. %
We see that our approach effectively and positively makes use of other representations. Philosophically, PropBank and FrameNet provide different granularities of semantic information; the complementary performance reflects these differences. Our results suggest that mid-tuning on different representations can provide alternative avenues for improvement, depending on the level of semantic granularity available or desired.

\begin{table}[t]
\resizebox{.99\columnwidth}{!}
{
\begin{tabular}{|c|p{7.1cm}|} 
\multicolumn{2}{c}{\textbf{Query:} The Statue of Liberty was built in 1875.} \\
\hline 
\multirow{3}{*}{\textbf{Bert-Base-Uncased}} &
\textbf{1st Match:} 10th Street Market was built in 1917 and expanded in 1926. (Score: 0.7529) \\
& \textbf{2nd Match:} As of 2013, a Daughters of the American Revolution plaque is present. (Score: 0.7473) \\
& \textbf{3rd Match:} At the entrance to the Trenton Battle Monument in Trenton New Jersey are the bronze statues of two soldiers. (Score: 0.7396) \\
\hdashline
\multirow{3}{*}{\textbf{Mid-Tuning}} & 
\textbf{1st Match:} The \underline{gatehouse} was \underline{constructed} in 1886 during the presidency of Grover Cleveland. (Score: 0.9962) \\
& \textbf{2nd Match:} The Art Deco \underline{building} was \underline{designed} by Thomas M. James in 1929. (Score: 0.9962) \\ 
& \textbf{3rd Match:} A \underline{large sculpture} was commissioned by U.S. President George Bush for the event. (Score: 0.9961) \\
\hline 
\end{tabular}
}
\caption{A KNN analysis from mid-tuning on a single Wikipedia file. Both baseline and mid-tuned models are compared against a given query and top-3 K nearest matches are presented.}
\label{KNN-2}
\end{table}

\subsection{KNN Experiments}
In this section, we perform a semantic search experiment to see the effect of mid-tuning alignment within a vector space. We embed all entries from a given set/corpus of sentences into a vector space. At search time, we select a query, embed it into the same vector space, and find the closest embeddings from the set using a K-Nearest-Neighbor search. For this experiment, we pick the BERT baseline model and compare the results with the best performing \textit{PropBank+Triplet} model. We use KNN as it is a simple, well-known algorithm that calculates the distance between two data points using cosine similarity.

We present two different examples of top-3 K-nearest matches against two random queries, where an entire wiki file with around 250k sentences is embedded into a vector space, and both BERT baseline and a mid-tuned model (PropBank+Triplet) are used to find the top nearest matches. In Table \ref{KNN-2}. For the query \textit{``The Statue of Liberty was built in 1875''}, all top matches with the mid-tuned models include some reference to either the keyword \textit{building} or \textit{built}, as represented by the underlined text in the table. Note also the third match for mid-tuning: ``sculpture'' is thematically relevant to ``building.'' We see that the mid-tuned model gives top-three matches which are more thematically related to the given query, as well as the notable difference between the cosine-similarity score of baseline model and mid-tuned models, which suggests that mid-tuning provides improved semantic alignment.


\section{Related Work}
\label{sec:related} 

%


Given a sentence, different meaning representation methods incorporate different semantic information. e.g. an Event or Frame that consists of a predicate (evokes the frame) and tells us what the event is about arguments and relations. Each representation method differs in how they handle predicate-argument relations, category of arguments (semantic roles), and the relation between events. \citet{zhang2019semantics} adds structured-specific semantic features (contextual semantic clues) alongside plain context-specific features and presents a modified language representation model named SemBERT, which improves accuracy over a large number of current SOTA models used for NLU and NLI tasks. In the past few years, various other semantic representation methods of text have also been introduced. %
UCCA~\cite{abend-rappoport-2013-universal} is a grammatical representation method to annotate the semantic distinctions within a sentence. UDS \cite{white-etal-2016-universal} adds cross-linguistic annotation protocols for Universal Dependency datasets.  %

Deep learning with triplet networks~\cite{hoffer2015deep} was inspired by the idea of Siamese networks where the network consists of multiple identical sub-networks. \Citet{ein2018learning} took the idea to learn thematic similarity between sentences by forming triplets and embed similar sentences within a section closer to each other, yielding better performance. \Citet{dor2018learning} use Triplet networks for semi-supervised NLP tasks and claims that the Triplet approach is very effective for semantic similarity prediction tasks. \Citet{reimers2019sentence} use Triplet loss objective function on large sections of Wikipedia sentences, and their results prove that learned embeddings from Triplet loss improve the generated sentence embeddings. %

Several methods have been introduced to encode information from text into vectors and using them for downstream tasks. Skip-Thought~\cite{kiros2015skip} uses encoder-decoder architecture to learn fixed-length representations of sentences using unsupervised learning w.r.t the order of sentences where they try to predict the surrounding. InferSent~\cite{conneau2017supervised} learn word embeddings using unsupervised learning by using labeled data from SNLI dataset \cite{bowman-etal-2015-large}. FastText~\cite{bojanowski2017enriching} learns embeddings by looking at sub-word information from the text using a Skip-gram model.

Several pre-fine-tuning approaches with BERT/Transformers have been introduced recently. \citet{phang2018sentence} introduce multi-stage fine-tuning where they fine-tune a BERT model on supervised task with labeled data. \citet{arase2019transfer} introduces an intermediate supervised training stage between pre-training and fine-tuning where they inject phrasal paraphrase relations into BERT.

\section{Conclusion}
We present a semantic mid-tuning approach that enhances the general understanding of a language encoder. We jointly encode semantic parses generated by FrameNet and SRL methods, which adds additional semantic knowledge to the encoders. We study both Classification and Triplet objective functions. This results in good improvement in a number of classification, semantic textual similarity, and inference tasks on GLUE, SuperGLUE, and SentEval benchmarks. 
Our work shows how semantic meaning can be effectively encoded from structured representations and transferred to non-structured encoders.

\paragraph{Acknowledgements and Funding Disclosure}
{\small
Some experiments were conducted on the UMBC HPCF, supported by the National Science Foundation under Grant No. CNS-1920079. %
This material is based in part upon work supported by the National Science Foundation under Grant Nos. IIS-1940931 and IIS-2024878. %
This material is also based on research that is in part supported by the Air Force Research Laboratory (AFRL), DARPA, for the KAIROS program under agreement number FA8750-19-2-1003. The U.S.Government is authorized to reproduce and distribute reprints for Governmental purposes notwithstanding any copyright notation thereon. The views and conclusions contained herein are those of the authors and should not be interpreted as necessarily representing the official policies or endorsements, either express or implied, of the Air Force Research Laboratory (AFRL), DARPA, or the U.S. Government.
}

\bibliography{aaai22}

\end{document}